\documentclass[runningheads]{llncs}

\usepackage{eccv}

\usepackage{eccvabbrv}

\usepackage{graphicx}
\usepackage{booktabs}
\usepackage{multirow}
\usepackage{pifont}
\newcommand{\cmark}{\ding{51}}
\newcommand{\xmark}{\ding{55}}
\usepackage[accsupp]{axessibility}  

\usepackage{floatrow}
\newfloatcommand{capbtabbox}{table}[][\FBwidth]

\usepackage{hyperref}

\usepackage{orcidlink}

\begin{document}

\title{Learning Dual-Level Deformable Implicit Representation for Real-World Scale Arbitrary Super-Resolution} 

\titlerunning{DDIR}

\author{Zhiheng Li\inst{1} \and
Muheng Li\inst{1} \and
Jixuan Fan\inst{2} \and Lei Chen\inst{1}\thanks{\;indicates the corresponding author.} \and Yansong Tang\inst{2} \and \\ Jiwen Lu\inst{1} \and Jie Zhou\inst{1}}

\authorrunning{Z. Li et al.}

\institute{Department of Automation, Tsinghua University, China \and 
Shenzhen International Graduate School, Tsinghua University, China \\
\email{\{lizhihan21, li-mh20, fjx23\}@mails.tsinghua.edu.cn, leichenthu@tsinghua.edu.cn, tang.yansong@sz.tsinghua.edu.cn,\\
\{lujiwen, jzhou\}@tsinghua.edu.cn}}

\maketitle

\begin{abstract}
  Scale arbitrary super-resolution based on implicit image function gains increasing popularity since it can better represent the visual world in a continuous manner. However, existing scale arbitrary works are trained and evaluated on simulated datasets, where low-resolution images are generated from their ground truths by the simplest bicubic downsampling. These models exhibit limited generalization to real-world scenarios due to the greater complexity of real-world degradations. To address this issue, we build a RealArbiSR dataset, a new real-world super-resolution benchmark with both integer and non-integer scaling factors fo the training and evaluation of real-world scale arbitrary super-resolution. Moreover, we propose a  Dual-level Deformable Implicit Representation (DDIR) to solve real-world scale arbitrary super-resolution. Specifically, we design the appearance embedding and deformation field to handle both image-level and pixel-level deformations caused by real-world degradations. The appearance embedding models the characteristics of low-resolution inputs to deal with photometric variations at different scales, and the pixel-based deformation field learns RGB differences which result from the deviations between the real-world and simulated degradations at arbitrary coordinates. Extensive experiments show our trained model achieves state-of-the-art performance on the RealArbiSR and RealSR benchmarks for real-world scale arbitrary super-resolution. The dataset and code are available at \url{https://github.com/nonozhizhiovo/RealArbiSR}. 
  \keywords{Real-World Scale Arbitrary Super-Resolution \and Deformable Implicit Neural Representation \and Appearance Embedding}
\end{abstract}

\section{Introduction}
\label{sec:intro}
Single image super-resolution (SISR) is a long-standing low-level task that reconstructs high-resolution (HR) images from their low-resolution (LR) inputs \cite{sr1,sr2,sr3,sr4,sr5,sr6,sr7,sr8,sr9}. It has been investigated for decades \cite{osr1,osr2,osr3,osr4,osr5,osr6,osr7,osr8,osr9}, and various sub-fields of SISR have been proposed \cite{sisr1, sisr2,sisr3,sisr4, blindsr1, blindsr2, blindsr3, blindsr4, blindsr5, blindsr6, blindsr7, realsr}. Among them, scale arbitrary super-resolution (SR) has been developed to generate HR images with arbitrary scales (even with non-integer ones) by only one model \cite{metasr, liif}. It better fulfills practical needs because we demand to zoom in and zoom out continuously in daily life. 

\begin{figure}[t]
  \centering
  \includegraphics[width=1\linewidth]{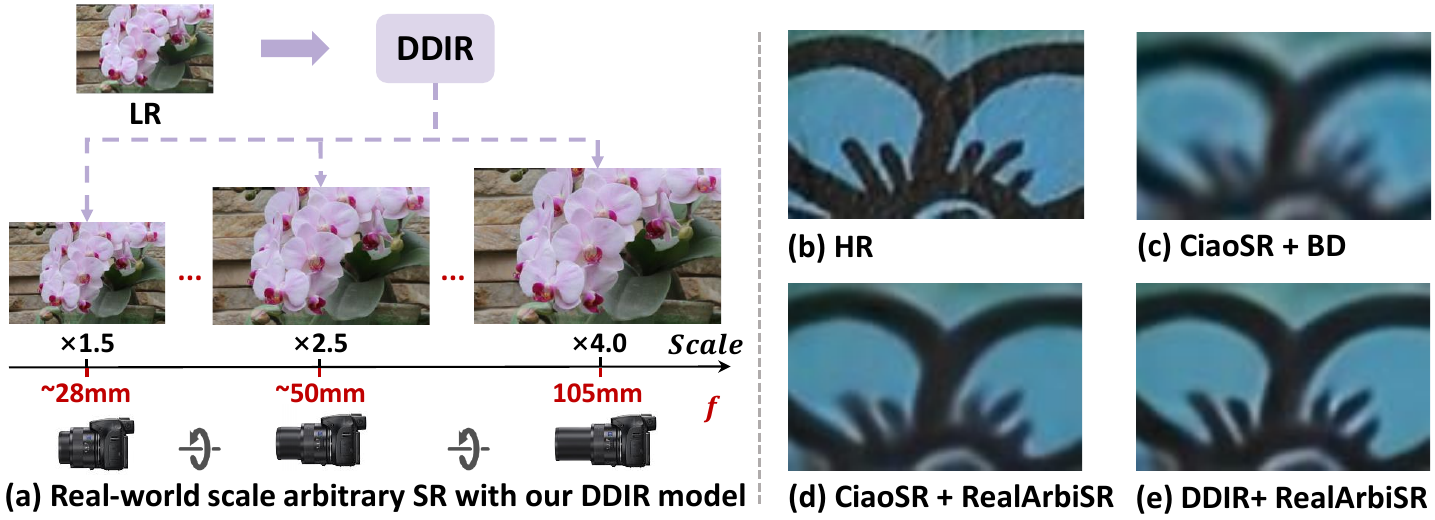}
  \caption{(a) We propose a Dual-level Deformable Implicit Representation (DDIR) to solve real-world scale arbitrary SR, simulating the continuous optical zoom of a DSLR camera by only one model. We compare (b) the HR image with the SR results ($\times 3.7$) of a real-world LR image generated by (c) CiaoSR\cite{ciaosr} trained on DIV2K dataset with bicubic degradation (CiaoSR+BD), (d) CiaoSR\cite{ciaosr} trained on RealArbiSR dataset with real-world degradation (CiaoSR+RealArbiSR), and (e) our DDIR model trained on RealArbiSR dataset with real-world degradation (DDIR+RealArbiSR). }
  \label{fig:1}
\end{figure}

Recent works in scale arbitrary SR are either based on convolutional neural networks (CNNs) \cite{metasr, arbsr} or implicit neural representations \cite{liif, lte, ipf}. However, these methods are trained and evaluated on simulated datasets, employing bicubic degradation only. By comparing Figure \ref{fig:1}(c) with Figure \ref{fig:1}(d), we can see the CiaoSR model \cite{ciaosr} trained on DIV2K dataset \cite{div2k} with bicubic downsampling is ineffective to reconstruct real-world HR images. Such synthetic degradation models cannot be generalized in real-world applications because real-world degradations are much more complex \cite{realsr, CDC}. To handle real-world degradation kernels, real-world SR has been investigated, and current works such as LP-KPN \cite{realsr} and CDC \cite{CDC} are based on CNNs and predict RGB values locally. However, existing real-world SR datasets are limited to integer scale factors (\eg, $\times 2$, $\times 3$, $\times 4$) and their models are constrained to work for one fixed scale factor.

To fill the gap between current SR research and the practical needs of zoom functionality, we construct a RealArbiSR dataset, the first real-world SR benchmark with both integer and non-integer scale factors for the training and evaluation of real-world scale arbitrary SR. To get the focal lengths of arbitrary scale factors, we use a checkerboard to calibrate the focal lengths for the desired scales, where the checkerboard is annotated with rectangles with changing areas for different scales. According to the calibrated focal lengths, we capture LR-HR image pairs, which are further aligned by an image registration algorithm \cite{realsr}. The RealArbiSR dataset provides a good benchmark for real-world scale arbitrary SR. It contains diverse indoor and outdoor scenes with both integer and non-integer scale factors. 

We propose a Dual-level Deformable Implicit Representation (DDIR) to solve the real-world scale arbitrary SR, simulating the continuous optical zoom of a high-end DSLR camera, as illustrated in Figure \ref{fig:1}(a). According to the analysis of real-world LR and HR images, we notice real-world degradation kernels can lead to image-level and pixel-level deformations on degraded images. We argue the image-level deformation is caused by photometric variations at different scales, and pixel-based deformation results from content-dependent and spatially variant degradation kernels. Therefore, we regard the problem of real-world scale arbitrary SR as a model that is deformed from the synthetic scale arbitrary SR (\eg, bicubic downsampling) along the channel dimension, and thus design a dual-level deformable implicit representation to learn image deformations at the image and pixel levels. For the image level, we model the characteristics of an LR image as the appearance embedding. The appearance embedding grants our model the ability to explain away photometric deformations between different scales, and improves the SR performance by a large margin. In addition, since real-world degradation kernels are content-dependent and spatially variant, we design a deformation branch to simulate the deformation field, which is calculated as the RGB differences that result from the deviations between the real-world and synthetic degradations at arbitrary spatial coordinates. The deformation field focuses on reconstructing image details in a continuous space at the pixel level.  Combining appearance embedding and deformation field, our DDIR model is capable of handling complex real-world degradation kernels and reconstructing real-world HR images with high fidelity. 

In summary, our contributions are threefold. (a) We build a RealArbiSR dataset, the first real-world SR dataset with both integer and non-integer scale factors in diverse scenes. The RealArbiSR dataset provides a good SR benchmark for the training and evaluation of real-world scale arbitrary SR. (b) We propose a dual-level deformable implicit representation to learn image-level and pixel-level deformations caused by complex real-world degradation kernels. (c) Experiments show our DDIR model achieves state-of-the-art performance on the RealArbiSR and RealSR benchmarks for real-world scale arbitrary SR. 

\section{Related Work}
\label{sec:relatedwork}
\textbf{Scale Arbitrary SR. } Scale arbitrary SR is first proposed by Meta-SR \cite{metasr}. Meta-SR utilizes the meta-upscale module to upscale LR inputs with arbitrary scales. ArbSR \cite{arbsr} proposes the scale-aware adaption blocks and a scale-aware upsampling layer. In addition to the CNN methods above, implicit neural representation \cite{inr1,inr2,inr3,inr4,inr5,inr6,inr7} has been widely used. LIIF \cite{liif} makes the first attempt by using the local implicit image function. Following LIIF, LTE \cite{lte} designs a local texture estimator to synthesize HR images in the Fourier domain. UltraSR \cite{ultrasr} introduces positional encoding with residual connections to the LIIF model to enhance the SR performance. OPE-SR \cite{opesr} proposes orthogonal position encoding for scale arbitrary SR. \cite{srno} designs super-resolution neural operator to learn the mapping between function spaces. CiaoSR \cite{ciaosr} proposes the continuous implicit attention-in-attention network for scale arbitrary SR. ITSRN \cite{ITSRN} designs an implicit transformer network to solve screen content SR with arbitrary scales. Different from the previous works, IPF \cite{ipf} introduces implicit pixel flow to generate perceptual-oriented HR results. However, current works of scale arbitrary SR all generate LR inputs by bicubic downsampling, which cannot simulate SR in real-world situations. \cite{realscalesr} solves real-world scale arbitrary SR by reverse modules SASRN and SARDN, but it is only trained and tested at the scale factor of $\times 2$.  

\noindent \textbf{Real-World SR Datasets. }Different from synthetic datasets, LR-HR image pairs in most real-world SR datasets are captured by adjusting the focal length of the camera. Qu et al. \cite{beamsplitter} use a beam splitter to collect LR-HR image pairs with two cameras. SupER dataset \cite{super} uses hardware binning to generate corresponding LR versions of ground truths. City100 dataset \cite{city100} contains 100 aligned image pairs that are captured from the printed postcards. Just like City100, D2CRealSR dataset \cite{d2c} takes photos of postcards in the laboratory environment to get image pairs with large scaling factors. SR-RAW dataset \cite{srraw} is the first real-world SR dataset captured in natural scenes, but their image pairs are not well aligned. RealSR dataset \cite{realsr} provides a good real-world SR benchmark by using an image registration algorithm to precisely align image pairs. Then, a large-scale DRealSR dataset \cite{CDC} is constructed. However, all the pixel-aligned real-world SR datasets captured in the indoor and outdoor environments \cite{realsr,CDC} only consist of image pairs with integer scale factors and thus are insufficient for the training and evaluation of real-world scale arbitrary SR. 

\noindent \textbf{Real-World SR Methods. }In contrast to the bicubic downsampling kernel, real-world degradation kernels are much more complicated because they are spatially variant and content-dependent. Zhang et al. \cite{srraw} propose a contextual bilateral loss for real-world SR. LP-KPN \cite{realsr} introduces a Laplacian pyramid network to learn spatially variant kernels and reconstruct HR images. CDC \cite{CDC} parses an image into three low-level components and proposes a component divide-and-conquer model to reconstruct HR images. DDet \cite{ddet} introduces a dual-path dynamic enhancement network. STCN \cite{stcn} designs a spatio-temporal correlation network and proposes a dual restriction to reduce the space of mapping functions in the real world. D2C-SR \cite{d2c} proposes a novel framework with divergence and convergence stages for real-world SR. These methods are based on CNN models, and only work for one specific integer scale factor. 

\section{The RealArbiSR Dataset}
\label{sec:dataset}

\subsection{Camera Calibration}
One significant feature of synthetic scale arbitrary SR is that it can predict HR images even at non-integer scale factors, where LR inputs can be easily generated by setting a bicubic scaling factor. Currently, real-world SR datasets only consist of LR-HR image pairs with multiple integer scale factors such as $\times 2$, $\times 3$, and $\times 4$ \cite{realsr,CDC}. Hence, no real-world dataset can be used to train and evaluate at non-integer scaling factors for real-world scale arbitrary SR. In the real-world situation, arbitrary scale factors especially non-integer ones are hard to indicate, because in general only a sparse set of focal length values are labeled on the zoom lens (\eg, Canon 24$\sim$105mm, $f$/4.0 zoom lens only displays the focal lengths of 105mm, 50mm, 35mm, and 24mm on the zoom ring). Also, the relation between the scaling factor and the focal length can be nonlinear. 

To get the desired focal lengths for arbitrary scaling factors, we use a checkerboard to calibrate the focal length of the zoom lens. The checkerboard is annotated with rectangles of various sizes for different scaling factors, as illustrated in the supplementary material. Considering the aberration effect is more severe with a wider angle of view, we take the images captured at the longest focal length as the ground truth of all scales, which corresponds to the smallest rectangle or equivalently the smallest field of view. In this way, the aberration effect can be minimized in all LR-HR image pairs. The widths and heights of other rectangles are enlarged by the desired scaling factors compared to the smallest ground-truth rectangle, so larger rectangles correspond to the LR inputs with larger scaling factors. During calibration, we first match the field of view of the longest focal length with the ground-truth rectangle by adjusting the camera position. After matching, we fix the camera at this steady position on a tripod. Then, we reduce the focal length to increase the field of view of the camera to match the larger rectangle of the desired scale factor. In this way, we can indicate and record the calibrated focal lengths for all scaling factors. For the training set, we collect the LR-HR image pairs of scaling factors from $\times 1.5$ to $\times 4$ with a step of $\times 0.5$ (including $\times 1.5$, $\times 2.0$, $\times 2.5$, $\times 3.0$, $\times 3.5$, and $\times 4.0$). For the test set, in addition to the scale factors that appeared in the training set (including $\times 1.5$, $\times 2.0$, $\times 2.5$, $\times 3.0$, $\times 3.5$, and $\times 4.0$), we further collected image pairs of the scale factors that are not present in the training set (\eg, $\times 1.7$, $\times 2.3$, $\times 2.7$, $\times 3.3$, and $\times 3.7$). We summarize the scale factors of our RealArbiSR dataset and compare with existing pixel-aligned real-world SR datasets captured in the indoor and outdoor environments in Table \ref{table1}.

\begin{table}[t]
\tabcolsep=0.29cm
\centering
\caption{Comparisons of the scale factors in the training set and test set between our RealArbiSR dataset and the existing pixel-aligned real-world SR datasets captured in the indoor and outdoor environments. }
\label{table1}
\begin{tabular}{l | cc }
\toprule
\multirow{2}{*}{Dataset} & \multicolumn{2}{c}{Scale Factor}  \\ 
& \multicolumn{1}{c}{Train set} & \multicolumn{1}{c}{Test set} \\ 
\midrule
RealSR\cite{realsr} & \multicolumn{1}{c}{$\times 2.0$, $\times 3.0$, $\times 4.0$} & \multicolumn{1}{c}{$\times 2.0$, $\times 3.0$, $\times 4.0$} \\ 
\midrule
DRealSR\cite{CDC} & \multicolumn{1}{c}{$\times 2.0$, $\times 3.0$, $\times 4.0$} & \multicolumn{1}{c}{$\times 2.0$, $\times 3.0$, $\times 4.0$}  \\ 
\midrule
RealArbiSR (Ours) & \multicolumn{1}{c}{\begin{tabular}[c]{@{}c@{}}$\times 1.5$, $\times 2.0$, $\times 2.5$, \\ $\times 3.0$, $\times 3.5$, $\times 4.0$\end{tabular}} & \begin{tabular}[c]{@{}c@{}}$\times 1.5$, $\times 1.7$, $\times 2.0$, $\times 2.3$, $\times 2.5$, $\times 2.7$, \\
$\times 3.0$, $\times 3.3$, $\times 3.5$, $\times 3.7$, $\times 4.0$\end{tabular} \\
\bottomrule
\end{tabular}
\end{table}

\subsection{Dataset Collection}
We use the DSLR camera (Canon 5D3) to capture LR-HR image pairs for dataset collection. The DSLR camera is equipped with a 24$\sim$105mm, $f$/4.0 zoom lens to cover the range of target scaling factors. We set the focal length of 105mm as the ground truth and the focal lengths of the LR inputs are indicated by the calibration procedure. When collecting images, the camera is fixed on a tripod. We first capture the ground-truth image at the focal length of 105mm, and then gradually zoom out the camera to collect LR inputs according to the calibrated focal lengths. We collect image pairs in diverse indoor and outdoor scenes to ensure our RealArbiSR dataset is generalized. We prefer to photograph objects at a distance of at least 3 meters. A large object distance can alleviate image deformations caused by aberrations. Also, we avoid photographing moving objects, since they are impossible to be aligned between image pairs. After collecting the ground truth and LR versions of all scale factors, we adopt the image registration algorithm \cite{realsr} to obtain pixel-wise aligned image pairs. For the scale factors from $\times 1.5$ to $\times 4.0$ with a step of $\times 0.5$, we get 250 scenes with 1500 LR-HR image pairs in total (Each scene has six image pairs for six scaling factors). 200 scenes are randomly chosen as the training set and the other 50 scenes are used as the test set. We further collect 83 scenes for the scale factors of $\times 1.7$, $\times 2.3$, $\times 2.7$, $\times 3.3$, and $\times 3.7$ as the test set. More details of the camera setting, image registration process, and the resolutions of the LR and HR versions at different scaling factors are discussed in the supplementary material. 

\section{Methods}
\label{sec:method}
\subsection{Analysis of Real-World Scale Arbitrary SR}
Before introducing our approach, we conduct a detailed analysis to compare the difference between the real-world scale arbitrary SR and the synthetic scale arbitrary SR. It better explains the motivation of our DDIR model. 

\begin{figure*}[t]
  \centering
  \includegraphics[width=1\linewidth]{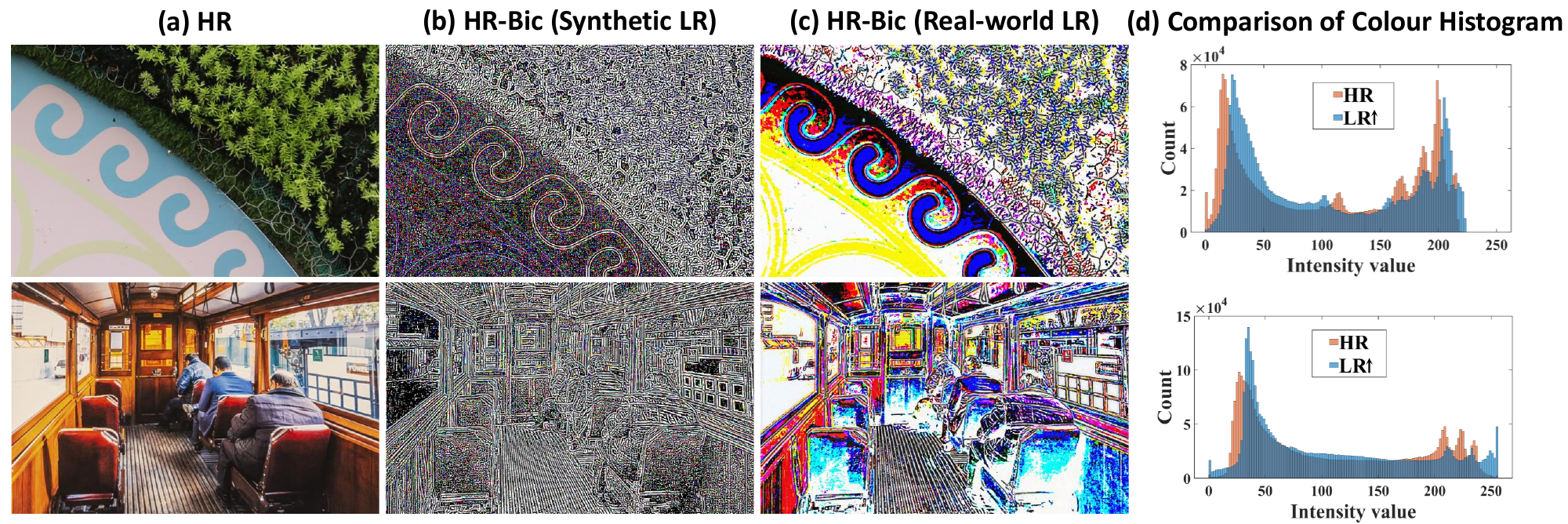}
  \caption{(a) The ground-truth images; (b) The images computed by subtracting the ground truths with their synthetic low-resolution versions which have been bicubically upscaled to the same resolution as the ground truth; (c) The images computed by subtracting the ground truths with their real-world low-resolution versions (bicubically upscaled); (d) The comparison of colour histograms between the ground truths and their real-world low-resolution versions (bicubically upscaled).}
  \label{fig2}
\end{figure*}

\noindent \textbf{Photometric Variation:} In real-world photographs, photometric variations on exposure, tone-mapping, white balance, etc., are unavoidable because the camera imaging pipeline is very complex. After changing the focal length of the camera, factors such as lighting conditions inside the field of view, camera sensors, image signal processing (ISP) pipeline, etc., can be all varied, leading to appearance variations over the whole image. To demonstrate such image-level variations, we compare the colour histogram of the ground-truth image with its bicubic-upscaled LR counterpart in Figure \ref{fig2}(d), illustrating the RGB values of LR images are generally shifted away from the ground-truth values. We argue these real-world photometric variations can be regarded as an image-level deformation from the template of synthetic degradation. Most existing real-world SR methods \cite{realsr, CDC} cannot solve this issue because they predict kernels and reconstruct high-resolution RGB targets locally. For the same reason, local implicit neural representation \cite{liif,lte,ipf} used in existing scale arbitrary SR works is insufficient to solve this real-world task since its receptive field is limited. 

\noindent \textbf{Pixel-based Deformation:} One difficulty of real-world scale arbitrary SR is that the model needs to predict real-world degradations at arbitrary spatial coordinates. To visualize and compare the effect of degradation kernels in real-world and synthetic scale arbitrary SR, we compute the RGB difference between the ground truth and bicubic-upscaled LR input in Figure \ref{fig2}. From Figure \ref{fig2}(b), we can see the RGB difference in synthetic scale arbitrary SR is only significant in high-frequency regions, including sharp edges and textures, with minor colour mismatching. In contrast, the RGB difference in real-world scale arbitrary SR is much more complex. In Figure \ref{fig2}(c), we find illumination and colour mismatches occur everywhere regardless of low-frequency or high-frequency regions in the real-world case. These mismatches are caused by content-dependent and spatially variant degradation kernels. 

According to the analysis above, we propose dual-level deformable implicit representation to address both image-level and pixel-level deformations in real-world scenarios. In Section \ref{subsec: global embedding}, we introduce an appearance embedding to address the deformation at the image level. In Section
\ref{subsec: local deformation field}, we design a deformation branch to model the deformation field and reconstruct image details at the pixel level. Figure \ref{fig3} shows the training pipeline of our DDIR model. 

\subsection{Overview}
\begin{figure*}[t]
  \centering
  \includegraphics[width=1\linewidth]{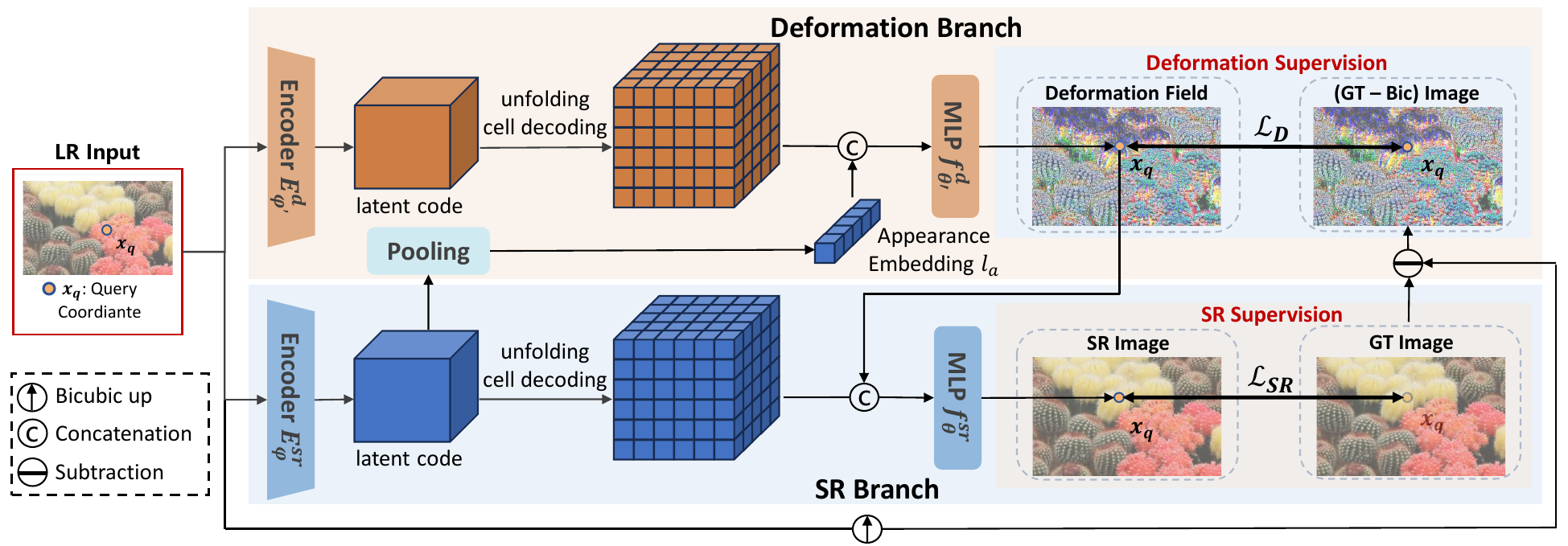}
  \caption{The training pipeline of our DDIR model. It consists of double branches, which are the deformation branch and the SR branch. Each branch is composed of an encoder and an MLP, taking the LR image and query coordinates as the inputs. The appearance embedding $l_a$ is computed as the spatial average pooling of the 2D feature map from the encoder $E_{\phi}^{sr}$ of the SR branch, which is fed into the decoding function $f_{\theta'}^{d}$ of the deformation branch by concatenation. The RGB output of the deformation branch is supervised by the deformation field. Then, the predicted deformation field feeds into the decoding function $f_{\theta}^{sr}$ of the SR branch by concatenation. Finally, the decoding function $f_{\theta}^{sr}$ of the SR branch outputs the target high-resolution RGB values at the query coordinates. Combining the appearance embedding and the deformation field, our DDIR model learns the dual-level deformable implicit representation to address the deformations at the image and pixel levels simultaneously. }
  \label{fig3}
\end{figure*}

The overall architecture of our DDIR model is illustrated in Figure \ref{fig3}. It is composed of two branches, which are the deformation branch and the SR branch. Each branch consists of one encoder and one decoding function, taking the pixel coordinate $(x, y)$ and the LR image as the inputs. Both decoding functions are parameterized by MLPs, and use local implicit neural representation to predict RGB values at query coordinates. Therefore, the prediction of the RGB values $I_q$ at an arbitrary query coordinate $x_q$ by a decoding function $f$ with trainable weights $\theta$ can be formulated as: 
\begin{equation}
  I_q = \sum_{i} \frac{S_i}{S}  \cdot f_{\theta}(m_{i}^{*}, x_q - x_{i}^{*}),
  \label{eq:1}
\end{equation}
where $m_{i}^{*}(i\in \{00, 01, 10, 11\})$ are the nearest latent code at the top-left, top-right, bottom-left, and bottom-right coordinates $x_{i}^{*}$ respectively, $S_i$ is the rectangle area between $x_{i}^{*}$ and $x_q$, and $S$ is the sum of all four $S_i$. Cell decoding and feature unfolding are also used.  

\subsection{Appearance Embedding}
\label{subsec: global embedding}
The receptive field of local implicit neural representation is limited. To adapt our DDIR model to variable photometric variations, we introduce an appearance embedding to the deformation branch to handle the image-level deformation. Here, the appearance embedding is simply taken as the spatial average pooling of the 2D feature map from the encoder $E_{\phi}^{sr}$ of the SR branch. Thus, the appearance embedding $l_a$ can be formulated as: 
\begin{equation}
  l_a = \frac{1}{WH}\sum_{x=1}^{W}\sum_{y=1}^{H} m_{x,y}^{*},
  \label{eq:2}
\end{equation}
where $m_{x,y}^*$ is the latent code at the coordinate of $x$ and $y$, and $W$ and $H$ are the width and height of the 2D feature map respectively. After getting the appearance embedding of the LR input, we concatenate it with the nearest latent code from the query coordinate $x_q$. With the appearance embedding, our DDIR model can `see' the characteristics of the whole LR input and is not purely local anymore. Although the appearance embedding is simply the spatial average pooling of the 2D feature map, we will show it can largely improve the metric results in experiments, especially in the real-world case.

\subsection{Deformation Field}
\label{subsec: local deformation field}
There is no way we can model the exact form of degradation kernels in real-world scale arbitrary SR because they are too complex to be known. Instead of directly predicting the form of degradation kernels, we simulate the effects of degradation kernels on RGB values. We regard bicubic downsampling as linear degradation and real-world downsampling as nonlinear degradation. We design the deformation branch to predict RGB differences that result from the derivations between real-world and synthetic degradation at arbitrary coordinates. More specifically, the RGB output of this branch is supervised by the RGB difference between the ground truth and the bicubic-upscaled LR input at this query point. We define the target of this branch as the deformation field $\Delta I(x_q)$ because it models the pixel-level deformation between the nonlinear and linear degradations at the arbitrary coordinate $x_q$, which can be computed as: 
\begin{equation}
  \Delta I(x_q)=I^{GT}(x_q)-I^{LR\uparrow}(x_q),
  \label{eq:3}
\end{equation}
where $I^{GT}(x_q)$ is the RGB value of the ground truth at the query coordinate $x_q$, and $I^{LR\uparrow}(x_q)$ is the RGB value of the upscaled LR (upscaled to the same size as the ground truth by bicubic) at the query coordinate $x_q$. By taking the residual between the ground truth and bicubic-upscaled LR input, the deformation field can simulate the spatially variant degradation effect and reconstruct texture details at the pixel level. Combining the appearance embedding and deformation field, our DDIR model learns a dual-level deformable implicit representation to address the image-level and pixel-level deformations simultaneously. 

\subsection{Network Architecture and Training}
The training pipeline of our DDIR model is shown in Figure \ref{fig3}. The network components and parameters of the two branches are separated. The deformation branch outputs the RGB values, supervised by the deformation field (Deformation Supervision). The SR branch outputs the target high-resolution RGB values, supervised by the ground truth (SR Supervision). In the deformation branch, the appearance embedding $l_a$ concatenates with the latent code from the deformation encoder $E_{\phi'}^{d}$ before feeding into the decoding function $f_{\theta'}^{d}$ of the deformation branch. In the SR branch, the predicted deformation field concatenates with the latent code from the SR encoder $E_{\phi}^{sr}$ before feeding into the decoding function $f_{\theta}^{sr}$ of the SR branch. The losses of the SR Supervision $\mathcal{L}_{SR}$ and the Deformation Supervision $\mathcal{L}_{D}$ are both L1 losses. Thus, the final loss $\mathcal{L}$ is formulated as the sum of these two losses:
\begin{equation}
  \mathcal{L} = \mathcal{L}_{SR}+\mathcal{L}_{D}.
  \label{eq:4}
\end{equation}
In inference, there is no computation of the bicubic-upscaled LR image and the subtraction between the ground-truth image and the bicubic-upscaled LR image, compared to the training pipeline. 

\section{Experiment}
\label{sec:exp}
\subsection{Experiment Setup}
We use the RealArbiSR dataset and the RealSR dataset \cite{realsr} for experiments of real-world scale arbitrary SR. The RealArbiSR dataset has 200 image pairs for training and either 50 or 83 image pairs for testing with various integer and non-integer scale factors. The RealSR dataset has around 400 image pairs for training and 100 image pairs for testing with the integer scale factors of $\times 2$, $\times 3$, and $\times 4$. We train the RealArbiSR dataset with the scale factors of $\times 1.5$, $\times 2.0$, $\times 2.5$, $\times 3.0$, $\times 3.5$, and $\times 4.0$, and test at the scale factors of $\times 1.5$, $\times 1.7$, $\times 2.0$, $\times 2.3$, $\times 2.5$, $\times 2.7$, $\times 3.0$, $\times 3.3$, $\times 3.5$, $\times 3.7$, and $\times 4.0$. The RealSR dataset is trained and tested with the scale factors of $\times 2$, $\times 3$, and $\times 4$. In the training time, we crop $48\times48$ patches as the inputs to the encoder. The corresponding HR patch with a random scale factor is also cropped as the ground-truth counterpart. 2304 pixels are randomly sampled from the ground-truth patch, and converted to coordinate-RGB pairs. We evaluate PSNR on the Y channel (\eg, luminance) of the transformed YCbCr space \cite{realsr, CDC}. 

The encoders $E_\phi^{sr}$ and $E_{\phi'}^{d}$ of both branches are either EDSR \cite{edsr} or RDN \cite{rdn} without the upsampling module. Both decoding functions $f_\theta^{sr}$ and $f_{\theta'}^{d}$ are 5-layer MLPs with ReLU activations and hidden dimensions of 256. we use an Adam \cite{adam} optimizer with an initial learning rate of $2\times10^{-4}$, which decays by 0.5 at every 200 epochs. The batch size is 16 and the models are trained for 1000 epochs. The last epoch is used for the final results. Experiments are conducted on two GeForce RTX 3090 GPUs. 

\subsection{Comparisons with State-of-the-Art}

\textbf{Quantitative Results. }In Table \ref{table2}, we compare the quantitative results between LIIF\cite{liif}, LTE\cite{lte}, CiaoSR\cite{ciaosr}, and our DDIR, using EDSR and RDN without the upsampling module as the encoders. Prior work \cite{realscalesr} is not included because its results are worse than the LIIF baseline. We can see our DDIR model achieves the best PSNR results at all the scale factors that appeared in the training set in the RealArbiSR dataset (including $\times 1.5$, $\times 2.0$, $\times 2.5$, $\times 3.0$, $\times 3.5$, and $\times 4.0$) and the RealSR dataset (\eg, $\times 2.0$, $\times 3.0$, and $\times 4.0$). In particular, compared with the previous SOTA method \cite{ciaosr}, our DDIR model achieves remarkable PSNR gains of 0.32dB under the RDN backbone ($\times 2.0)$ on the RealArbiSR dataset. Even for the scale factors that are not present in the training set (such as $\times 1.7$, $\times 2.3$, $\times 2.7$, $\times 3.3$, and $\times 3.7$), our DDIR model also achieves the best PSNR results at all these scale factors, as illustrated in Table \ref{table3}. These experimental results show the appearance embedding and the deformation field handle real-world degradation kernels from the perspectives of image-level and pixel-level deformations properly, resulting in robust metric gains at all the scale factors under both backbones in the RealArbiSR and RealSR datasets. 

We further conduct the cross-dataset testing in Table \ref{table4}. We train the models on the RealArbiSR dataset and test them on the RealSR dataset. Table \ref{table4} shows our DDIR model achieves the best metric results in the cross-dataset experiment. 

\begin{table}[t]
  \caption{Quantitative comparison on RealArbiSR and RealSR datasets in PSNR(dB). The highest PSNR at each scale factor of each dataset is bolded. One model is trained and tested at the scale factors from $\times 1.5$ to $\times 4.0$ with a step of $\times 0.5$ in RealArbiSR dataset. Another model is trained and tested at the integer scale factors including $\times 2.0$, $\times 3.0$, and $\times 4.0$ in RealSR dataset. }
  \label{table2}
  \centering
  \begin{tabular}{ l |cccccc | ccc}
    \toprule
     \multirow{2}{*}{Method} & \multicolumn{6}{c|}{RealArbiSR} & \multicolumn{3}{c}{RealSR} \\ 
     & $\times 1.5$ & $\times 2.0$ & $\times 2.5$ & $\times 3.0$ & $\times 3.5$ & $\times 4.0$ & $\times 2.0$ & $\times 3.0$ & $\times 4.0$ \\
    \midrule
      Bicubic\cite{realsr} & 35.46 & 32.45 & 30.69 & 29.42 & 28.50 & 27.80 & 31.67 & 28.61 & 27.24 \\ 
    \midrule
    
     EDSR-baseline\cite{edsr} & - & 34.26 & - & 31.12 & - & 29.47 & 33.88 & 30.86 & 29.09 \\
    
     EDSR-LIIF \cite{liif} & 37.14 & 34.37 & 32.54 & 31.28 & 30.29 & 29.63& 33.87 & 30.77 & 29.18 \\

     EDSR-LTE\cite{lte} & 37.16 & 34.34 & 32.53 & 31.26 & 30.30 & 29.68 & 33.94 & 30.80 & 29.21 \\

     EDSR-CiaoSR\cite{ciaosr} & 37.23 & 34.54 & 32.80 & 31.52 & 30.57 & 29.88 & 34.08 & 30.97 & 29.37 \\
    
     EDSR-DDIR  & \textbf{37.51} & \textbf{34.85} & \textbf{33.02} & \textbf{31.78} & \textbf{30.80} & \textbf{30.05} & \textbf{34.19} & \textbf{31.02} & \textbf{29.39} \\
    \midrule
     RDN-LIIF \cite{liif} & 37.14 & 34.41 & 32.60 & 31.40 & 30.34 & 29.70 & 33.99 & 30.90 & 29.29 \\

     RDN-LTE\cite{lte} & 37.24 & 34.52 & 32.76 & 31.53 & 30.54 & 29.84 & 34.01 & 30.93 & 29.29 \\

     RDN-CiaoSR\cite{ciaosr} & 37.38 & 34.70 & 32.96 & 31.68 & 30.77 & 30.07 & 34.26 & 31.14 & 29.45 \\
    
     RDN-DDIR & \textbf{37.63} & \textbf{35.02} & \textbf{33.20} & \textbf{31.91} & \textbf{30.94} & \textbf{30.21} & \textbf{34.35} & \textbf{31.15} & \textbf{29.48}\\
    \bottomrule
  \end{tabular}
\end{table}

\begin{table}[t!]
  \caption{Quantitative comparison on RealArbiSR in PSNR(dB). The highest PSNR at each scale factor of each dataset is bolded. One model is trained at the scale factors from $\times 1.5$ to $\times 4.0$ with a step of $\times 0.5$, but tested at the scale factors of $\times 1.7$, $\times 2.3$, $\times 2.7$, $\times 3.3$, and $\times 3.7$ in RealArbiSR dataset.}
  \label{table3}
  \centering
  \begin{tabular}{ l |ccccc | ccccc}
    \toprule
    \multirow{2}{*}{Method} & \multicolumn{5}{c|}{EDSR Backbone} & \multicolumn{5}{c}{RDN Backbone} \\ 
    & $\times 1.7$ & $\times 2.3$ & $\times 2.7$ & $\times 3.3$ & $\times 3.7$ & $\times 1.7$ & $\times 2.3$ & $\times 2.7$ & $\times 3.3$ & $\times 3.7$ \\
    \midrule
     Bicubic & 33.53 & 31.05 & 30.12 & 29.03 & 28.48 & 33.53 & 31.05 & 30.12 & 29.03 & 28.48 \\ 
    \midrule
    
     LIIF \cite{liif} & 34.63 & 32.33 & 31.39 & 30.22 & 29.64 & 34.66 & 32.40 & 31.45 & 30.28 & 29.71   \\

     LTE\cite{lte} & 34.65 & 32.29 & 31.30 & 30.10 & 29.51 & 34.74 & 32.44 & 31.55 & 30.39 & 29.81 \\

     CiaoSR\cite{ciaosr} & 34.49 & 32.44 & 31.64 & 30.48 & 29.87 & 34.54 & 32.50 & 31.67 & 30.56 & 29.96 \\
     DDIR & \textbf{34.90} & \textbf{32.73} & \textbf{31.80} & \textbf{30.61} & \textbf{30.01} & \textbf{35.07} & \textbf{32.88} & \textbf{31.96} & \textbf{30.75} & \textbf{30.15} \\
    \bottomrule
  \end{tabular}
\end{table}

\begin{table}
  \caption{Quantitative comparison of the cross-dataset testing in PSNR(dB). The highest PSNR at each scale factor is bolded. One model is trained at the scale factors from $\times 1.5$ to $\times 4.0$ with a step of $\times 0.5$ in RealArbiSR dataset, and tested at the scale factors of $\times 2.0$, $\times 3.0$, and $\times 4.0$ in RealSR dataset.}
  \label{table4}
  \centering
  \begin{tabular}{ l |ccc|ccc}
    \toprule
    \multirow{2}{*}{Method} & \multicolumn{3}{c|}{EDSR Backbone} & \multicolumn{3}{c}{RDN Backbone} \\ 
     & $\times 2.0$ & $\times 3.0$ & $\times 4.0$ & $\times 2.0$ & $\times 3.0$ & $\times 4.0$ \\
    \midrule
     LIIF \cite{liif} & 32.47 & 29.54 & 28.03  & 32.39 & 29.54 & 28.02 \\

     LTE\cite{lte} & 32.48 & 29.52 & 28.06 & 32.27 & 29.47 & 28.06  \\

     CiaoSR\cite{ciaosr} & 32.38 & 29.56 & 28.19  & 32.42 & 29.58 & 28.17 \\
     DDIR & \textbf{32.58} & \textbf{29.72} & \textbf{28.25}  & \textbf{32.59} & \textbf{29.70} & \textbf{28.18} \\

    \bottomrule
  \end{tabular}
\end{table}

\begin{figure*}[t!]
  \centering
  \includegraphics[width=1\linewidth]{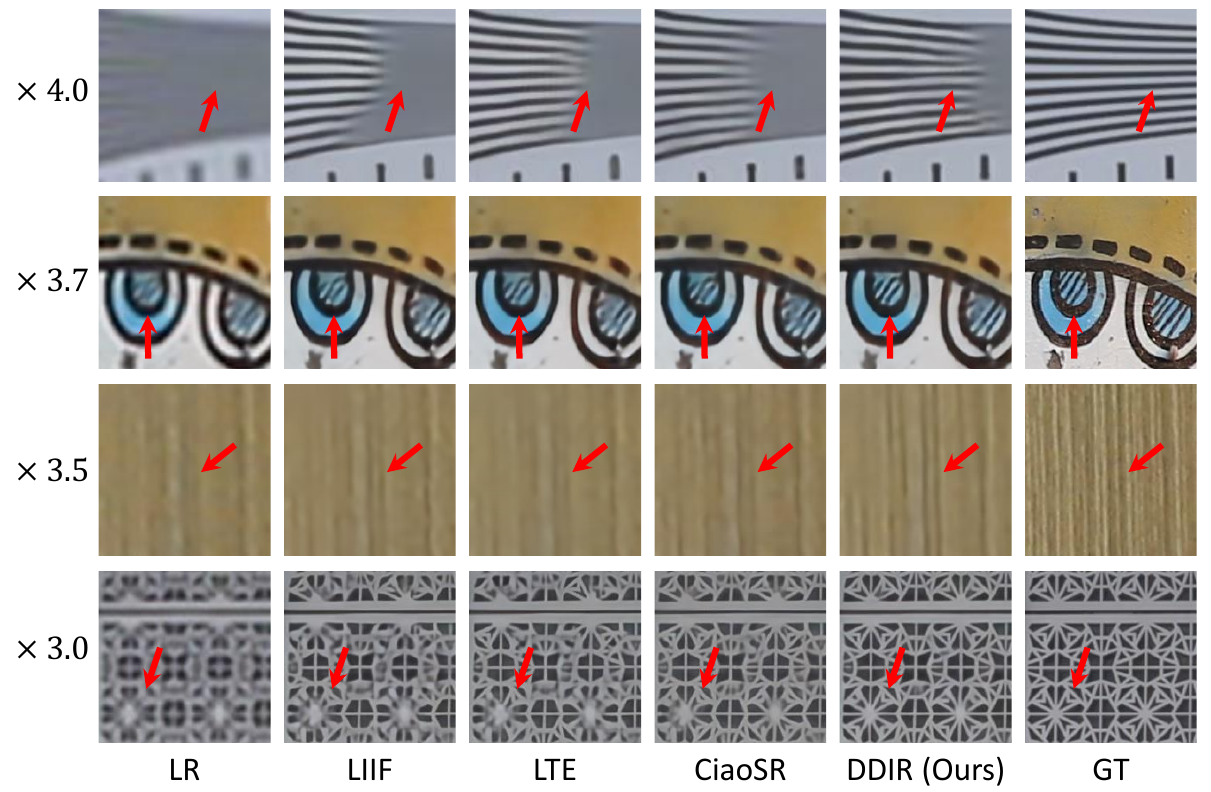}
  \caption{Qualitative comparisons between different methods on benchmarks. Zoom in to have better views.}
  \label{fig4}
\end{figure*}

\noindent \textbf{Qualitative Results. }We present a qualitative comparison between LIIF \cite{liif}, LTE\cite{lte}, CiaoSR\cite{ciaosr} and our DDIR in Figure \ref{fig4}. It shows our DDIR model obtains better visual quality than the competitors, reconstructing sharper edges and more natural details. In contrast, the results of other methods \cite{liif, lte, ciaosr} suffer from unpleasant details, especially blurry edges. Taking the first group of the images as an example, our DDIR can reconstruct more lines with sharp edges. However, more of the lines in other methods are blurred. These qualitative comparisons prove our DDIR reconstructs HR images with better texture details due to the use of appearance embedding and deformation field. 

\subsection{Analysis of Scale Factors in RealArbiSR Dataset}
Compared with the existing real-world SR dataset (\eg, RealSR\cite{realsr} and DRealSR dataset\cite{CDC}), our RealArbiSR dataset has three more non-integer scale factors including $\times 1.5$, $\times 2.5$, and $\times 3.5$ in the training set. To demonstrate our RealArbiSR dataset is more suitable for the training of real-world scale arbitrary SR due to the presence of these non-integer scale factors, we compare the metric results of the models trained with only integer scale factors and with all scale factors in Table \ref{table5}. We can see the models which are trained at all scale factors (including $\times 1.5$, $\times 2.0$, $\times 2.5$, $\times 3.0$, $\times 3.5$, and $\times 4.0$, indicated as `All' in Table \ref{table5}) perform better than the ones trained only at integer scale factors (including $\times 2.0$, $\times 3.0$, and $\times 4.0$, indicated as `$\times 2$$\times 3$$\times 4$' in Table \ref{table5}). Further experimental results are presented in the supplemental material. 

\begin{table*}[t]
  \caption{Quantitative Analysis of training scale factors in RealArbiSR dataset. The highest PSNR at each scale factor on each method is bolded. `$\times 2$$\times 3$$\times 4$' represents the models are trained at the scale factors of $\times 2.0$, $\times 3.0$, and $\times 4.0$. `All' represents the models are trained at the scale factors of  $\times 1.5$, $\times 2.0$, $\times 2.5$, $\times 3.0$, $\times 3.5$, and $\times 4.0$. }
  \label{table5}
  \centering
  \begin{tabular}{c | c|cccccc }
    \toprule
    Method & Training Scale &  $\times 1.5$ & $\times 2.0$ & $\times 2.5$ & $\times 3.0$ & $\times 3.5$ & $\times 4.0$  \\
    \midrule
    \multirow{2}{*}{EDSR-LIIF\cite{liif}} & $\times 2$$\times 3$$\times 4$ & 36.70 & 34.20 & 32.39 & 31.19 & 30.22 & 29.59  \\
    & All & \textbf{37.14} & \textbf{34.37} & \textbf{32.54} & \textbf{31.28} & \textbf{30.29} & \textbf{29.63}  \\
    \midrule
    \multirow{2}{*}{EDSR-LTE\cite{lte}} & $\times 2$$\times 3$$\times 4$ & 36.89 & 34.23 & 32.40 & 31.18 & 30.21 & 29.59  \\
    & All & \textbf{37.16} & \textbf{34.34} & \textbf{32.53} & \textbf{31.26} & \textbf{30.29} & \textbf{29.68}  \\\midrule
    \multirow{2}{*}{EDSR-CiaoSR\cite{ciaosr}} & $\times 2$$\times 3$$\times 4$ & 36.85 & 34.45 & 32.69 & 31.45 & 30.49 & 29.82  \\
    & All & \textbf{37.23} & \textbf{34.54} & \textbf{32.80} & \textbf{31.52} & \textbf{30.57} & \textbf{29.88}  \\\midrule
    \multirow{2}{*}{EDSR-DDIR} & $\times 2$$\times 3$$\times 4$ & 37.21 & 34.63 & 32.80 & 31.61 & 30.61 & 29.90  \\
    & All & \textbf{37.51} & \textbf{34.85} & \textbf{33.02} & \textbf{31.78} & \textbf{30.80} & \textbf{30.05}  \\
    \bottomrule
  \end{tabular}
\end{table*}
\subsection{Ablation Study}
We show an ablation study in Table \ref{table6} to demonstrate the effect of the appearance embedding and deformation field. After removing appearance embedding, the PSNR results are reduced at all scale factors. To illustrate the effect of deformation field, we remove the deformation field and the branch, concatenating the appearance embedding with the latent code of the SR branch. In this case, the PSNR results also drop at all scale factors. Without the appearance embedding and deformation field, the model performs the worst at all these scale factors. 

\begin{table}[t]
  \tabcolsep=0.15cm
  \caption{Quantitative ablation study of EDSR-DDIR on RealArbiSR dataset in PSNR(dB). The highest PSNR at each scale factor is bolded. }
  \label{table6}
  \centering
  \begin{tabular}{cc|cccccc}
    \toprule
    Deformation & Appearance & \multicolumn{6}{c}{Scale} \\
    Field & Embedding & $\times 1.5$ & $\times 2.0$ & $\times 2.5$ & $\times 3.0$ & $\times 3.5$ & $\times 4.0$ \\
    \hline
    \xmark & \xmark & 37.09 & 34.32 & 32.52 & 31.29 & 30.33 & 29.69 \\
    \cmark & \xmark & 37.26 & 34.50 & 32.66 & 31.41 & 30.41 & 29.72 \\
    \xmark & \cmark & 37.32 & 34.69 & 32.86 & 31.61 & 30.64 & 29.90 \\
    \cmark & \cmark & \textbf{37.51} & \textbf{34.85} & \textbf{33.02} & \textbf{31.78} & \textbf{30.80} & \textbf{30.05} \\
    \bottomrule
    \end{tabular}
\end{table}

\subsection{Analysis of Appearance Embedding on Real-World SR}
To demonstrate appearance embedding is particularly useful in real-world scale arbitrary SR, we compare the experimental results with and without the appearance embedding in the synthetic (\eg, bicubic downsampling) and the real-world scale arbitrary SR. Specifically, we use the EDSR-LIIF \cite{liif} model as the baseline and choose to concatenate or not concatenate the appearance embedding with the latent code before feeding into the MLP. We train and evaluate these models on DIV2K dataset (bicubic degradation) \cite{div2k} and our RealArbiSR dataset. In Table \ref{table7}, we can see the PSNR gains by adding the appearance embedding in the real-world dataset are significantly higher than the gains in the synthetic dataset. This proves the appearance embedding is much more useful in the real-world case. Since bicubic downsampling cannot have image-level deformations, we argue there exist image-level deformations in real-world SR. By adding appearance embedding, our DDIR model can handle image-level degradations caused by photometric variations in real-world photographs, leading to a more significant improvement in the real-world scenario than the synthetic one. 
\begin{table*}[t]
  \caption{Quantitative comparison of appearance embedding in synthetic and real-world scale arbitrary SR in PSNR(dB) on RGB channels. EDSR-LIIF(+a) refers to adding appearance embedding to the EDSR-LIIF baseline. The highest PSNR at each scale factor is bolded. The models are trained and evaluated on RealArbiSR and DIV2K datasets. RealArbiSR dataset is tested at the scale factors from $\times 1.5$ to $\times 4.0$ with a step of $\times 0.5$ and DIV2K dataset is tested at the scale factors of $\times 2.0$, $\times 3.0$, and $\times 4.0$.}
  \label{table7}
  \centering
  \begin{tabular}{c |cccccc | ccc}
    \toprule
    \multirow{2}{*}{Method} & \multicolumn{6}{c|}{RealArbiSR} & \multicolumn{3}{c}{DIV2K} \\ 
    & $\times 1.5$ & $\times 2.0$ & $\times 2.5$ & $\times 3.0$ & $\times 3.5$ & $\times 4.0$ & $\times 2.0$ & $\times 3.0$ & $\times 4.0$ \\
    \midrule
    EDSR-LIIF \cite{liif} & 34.88 & 32.27 & 30.49 & 29.27 & 28.29 & 27.62 & 34.67 & 30.96 & 29.00 \\
    EDSR-LIIF(+a) & \textbf{35.09} & \textbf{32.64} & \textbf{30.88} & \textbf{29.64} & \textbf{28.68} & \textbf{27.96} & \textbf{34.74} & \textbf{31.05} & \textbf{29.07} \\
    \bottomrule
  \end{tabular}
\end{table*}

\section{Conclusion}
\label{sec:conclusion}

In this work, we contribute a RealArbiSR dataset, the first real-world SR benchmark with both integer and non-integer scale factors in diverse scenes for the training and evaluation of real-world scale arbitrary SR. We propose dual-level deformable implicit representation to solve this problem. Specifically, the appearance embedding and deformation field are designed to handle image-level and pixel-level deformations caused by real-world degradation kernels. Extensive experiments show our DDIR model is capable of dealing with complex real-world degradations and reconstructing real-world HR images with high fidelity, achieving state-of-the-art performance on both RealArbiSR and RealSR benchmarks. As for limitations, our RealArbiSR dataset uses one camera to collect image pairs. The differences in imaging pipelines among cameras can lead to more diverse degradations. In the future, we will build a large-scale dataset by using more DSLR cameras to cover diverse real-world degradation among devices.

\noindent \textbf{Acknowledgement. }This work was supported in part by the National Natural Science Foundation of China under Grant 62321005, Grant 62336004, Grant 62125603, and Grant 62306031.

%
%
\bibliographystyle{splncs04}
\bibliography{main}

\newpage

\section*{Supplementary Material}
\appendix

\section{Further Details about RealArbiSR Dataset}
\begin{figure*}
  \centering
  \includegraphics[width=0.9\linewidth]{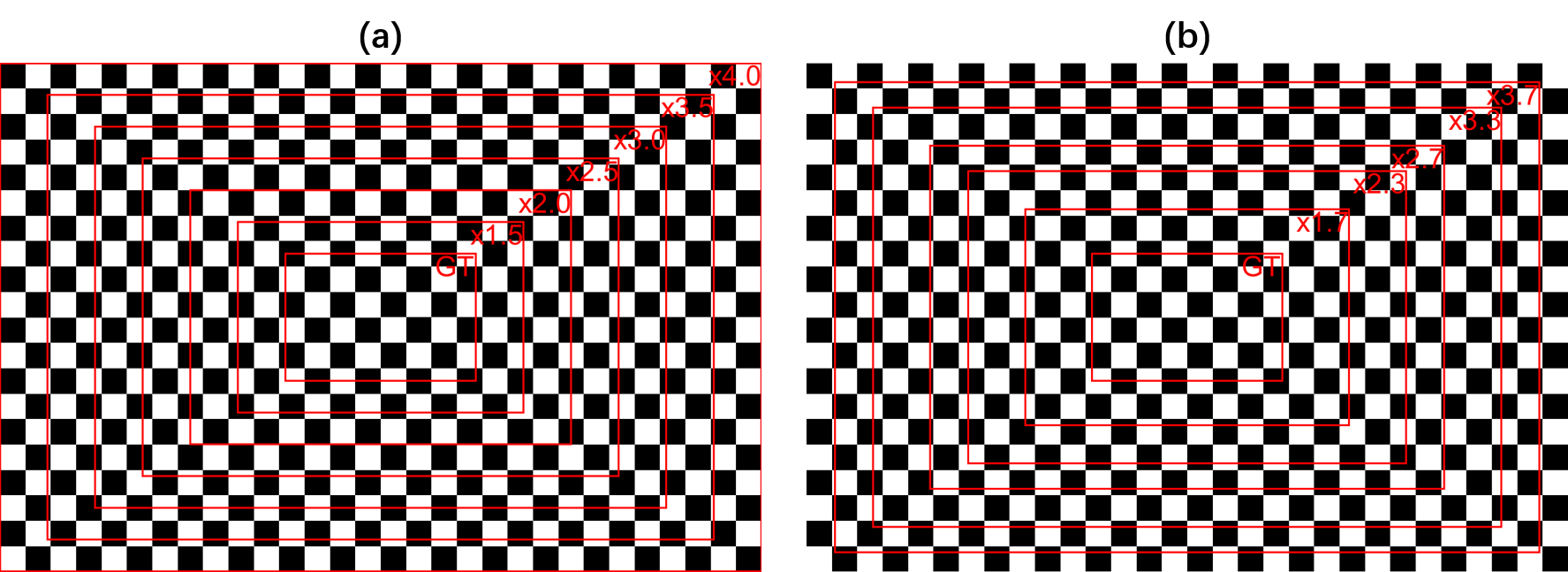}
  \caption{The checkerboards for the calibration of the focal lengths with the scale factors of (a) $\times$1.5, $\times$2.0, $\times$2.5, $\times$3.0, $\times$3.5, $\times$4.0; and (b) $\times$1.7, $\times$2.3, $\times$2.7, $\times$3.3, $\times$3.7. } 
  \label{figS1}
\end{figure*}

\begin{figure*}
  \centering
  \includegraphics[width=0.67\linewidth]{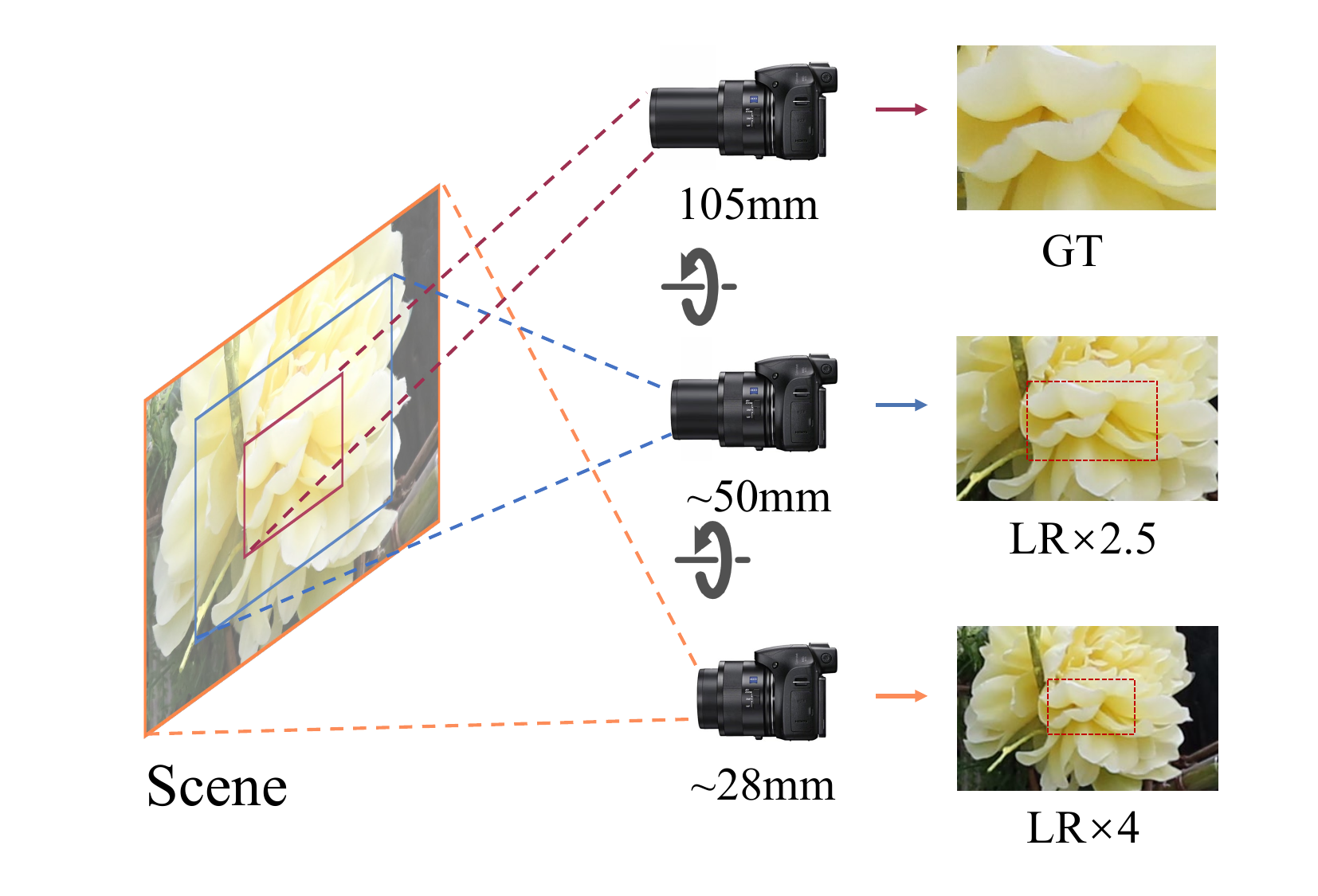}
  \caption{The illustration of dataset collection. The images taken at the longest focal length (105mm) are used as the ground truths, and the low-resolution versions are cropped from the red-dotted regions from the images taken at shorter focal lengths. } 
  \label{figS2}
\end{figure*}

\textbf{Camera Calibration and Camera Setting. } We use two checkerboards for the calibration of the focal lengths in RealArbiSR dataset. The first checkerboard is designed for the scale factors of $\times 1.5$, $\times 2.0$, $\times 2.5$, $\times 3.0$, $\times 3.5$, and $\times 4.0$. In this case, the checkerboard is annotated with seven concentric rectangles, which are labeled by `GT', `$\times 1.5$', `$\times 2.0$', `$\times 2.5$', `$\times 3.0$', `$\times 3.5$', and `$\times 4.0$' at the top-right corners of the corresponding rectangles, as illustrated in Figure \ref{figS1}(a). The second checkerboard is designed for the rest scale factors, annotated with six rectangles with the labels of `GT', `$\times 1.7$', `$\times 2.3$', `$\times 2.7$', `$\times 3.3$', and `$\times 3.7$', demonstrated in Figure \ref{figS1}(b). The camera is set to aperture priority mode. The focus, exposure, white balance, and ISO are set to automatic. We prefer to capture images in bright light conditions, because captured images tend to be noisy in the dark environment. We make sure there is no inappropriate blur due to depth-of-field by manual check. While collecting images, we gradually zoom out the camera to collect all LR-HR image pairs. For each scene, we take the images captured at the longest focal length as the ground truths, and the low-resolution versions are cropped from the red-dotted regions at shorter focal lengths, as shown in Figure \ref{figS2}.  

\noindent \textbf{Image Alignment. }We first use the image registration algorithm with luminance adjustment \cite{realsr} to coarsely align the low-resolution images with their high-resolution ground truths, and then crop the corresponding central regions of all images. In this way, the aberration effect can be minimized in the LR-HR image pairs. Next, we finer align the cropped LR-HR image pairs by adopting the image registration algorithm with luminance adjustment again. The borders of the aligned images are shaved to remove the distorted regions caused by the registration algorithm. We set 5 iterations for the optimization process of both coarse and fine alignments. After all these image pre-processing, we conduct a careful manual check for all images. Image pairs with inappropriate blur, moving objects, inappropriate exposure, etc., are all discarded. 

\begin{figure}
  \centering
  \includegraphics[width=0.8\linewidth]{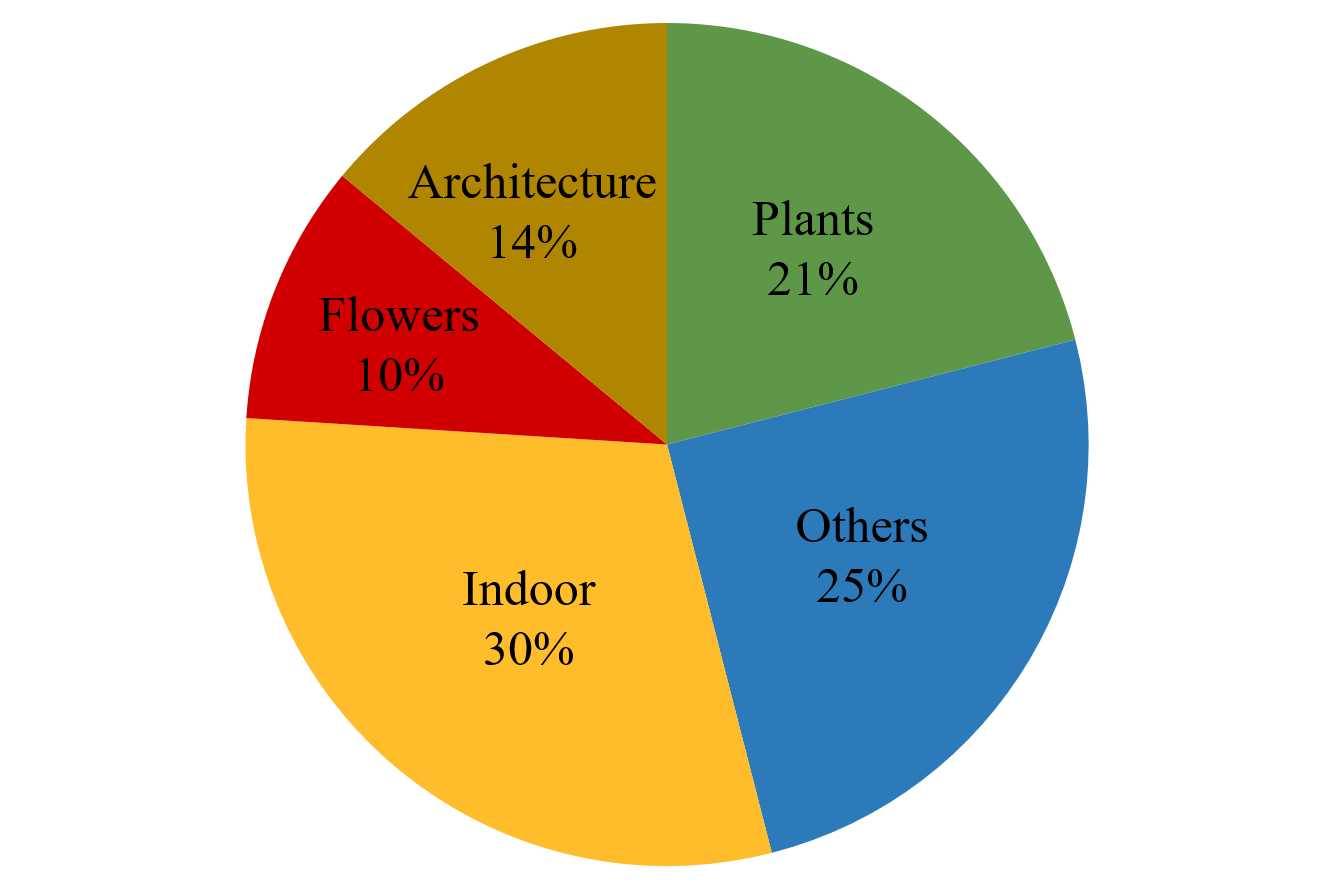}
  \caption{The content distribution of our RealArbiSR dataset. } 
  \label{figSpie}
\end{figure}

\begin{figure*}
  \centering
  \includegraphics[width=\linewidth]{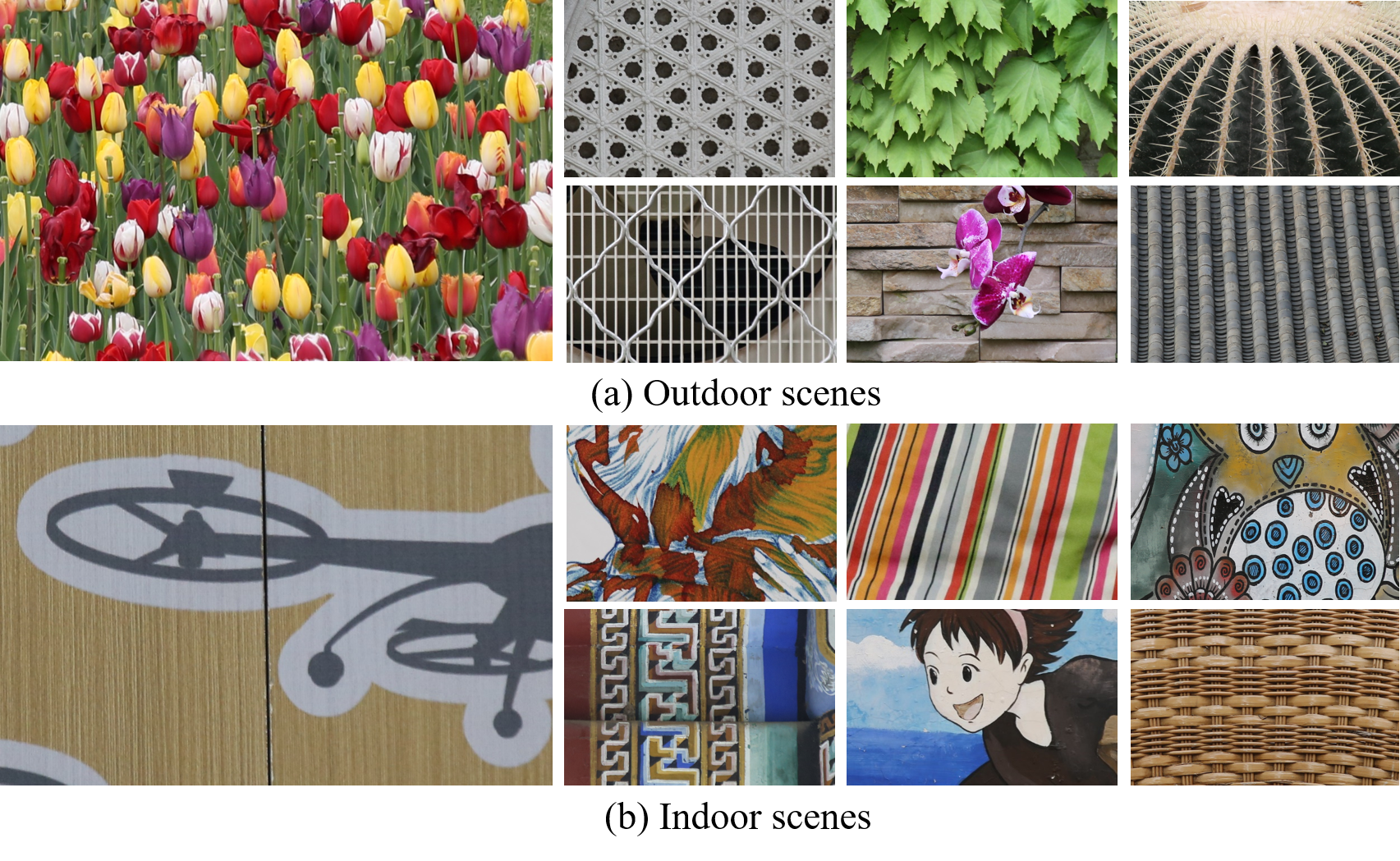}
  \caption{Image examples of the RealArbiSR dataset. } 
  \label{figS3}
\end{figure*}

\noindent \textbf{Dataset Statistics. } The resolutions of LR and HR images for different scale factors are listed in Table \ref{tableS1}. The RealArbiSR dataset covers diverse scenes in indoor and outdoor environments. We present the content distribution of our RealArbiSR dataset in Figure \ref{figSpie}. Some ground-truth examples of the RealArbiSR dataset are illustrated in Figure \ref{figS3}. 

\begin{table*}[t]
  \caption{The resolutions of LR and HR images for different scale factors in RealArbiSR dataset. }
  \label{tableS1}
  \centering
  \begin{tabular}{c|cccccc}
    \toprule
    Scale factor & $\times 1.5$ & $\times 2.0$ & $\times 2.5$ & $\times 3.0$ & $\times 3.5$ & $\times 4.0$  \\
    \hline
    HR & $1212\times792$ & $1196\times776$ & $1180\times760$ & $1164\times744$ & $1148\times728$ & $1132\times712$  \\ 
    LR & $808\times528$ & $598\times388$ & $472\times304$ & $388\times248$ & $328\times208$ & $283\times178$  \\

    \hline 

    Scale factor & $\times 1.7$ & $\times 2.3$ & $\times 2.7$ & $\times 3.3$ & $\times 3.7$ &   \\
    \hline
    HR & $1241\times816$ & $1219\times782$ & $1215\times783$ & $1188\times792$ & $1184\times777$   \\ 
    LR & $730\times480$ & $530\times340$ & $450\times290$ & $360\times240$ & $320\times210$ &  \\
    
    \bottomrule
  \end{tabular}
\end{table*}

\section{Further Analysis of Training Scaling Factors in RealArbiSR Dataset}
To further analyze the effect of non-integer scale factors in the training of real-world scale arbitrary SR, we present the experimental results with the RDN \cite{rdn} backbone at all scale factors in Table \ref{tableS2} and \ref{tableS3}. As shown in Table \ref{tableS2} and \ref{tableS3}, the models which are trained at all scale factors (including $\times 1.5$, $\times 2.0$, $\times 2.5$, $\times 3.0$, $\times 3.5$, and $\times 4.0$, indicated as `All' in Table \ref{tableS2} and \ref{tableS3}) perform better than the ones trained only at integer scale factors (including $\times 2.0$, $\times 3.0$, and $\times 4.0$, indicated as `$\times 2$$\times 3$$\times 4$' in Table \ref{tableS2} and \ref{tableS3}).

\begin{table*}
  \caption{Quantitative Analysis of training scale factors in RealArbiSR dataset. The highest PSNR at each scale factor on each method is bolded. `$\times 2$$\times 3$$\times 4$' represents the models are trained at the scale factors of $\times 2.0$, $\times 3.0$, and $\times 4.0$. `All' represents the models are trained at the scale factors of  $\times 1.5$, $\times 2.0$, $\times 2.5$, $\times 3.0$, $\times 3.5$, and $\times 4.0$. The models are tested at the scale factors from $\times 1.5$ to $\times 4.0$ with a step of $\times 0.5$ in RealArbiSR dataset. }
  \label{tableS2}
  \centering
  \begin{tabular}{c | c|cccccc }
    \toprule
    Method & Training Scale &  $\times 1.5$ & $\times 2.0$ & $\times 2.5$ & $\times 3.0$ & $\times 3.5$ & $\times 4.0$  \\
    \midrule
    \multirow{2}{*}{RDN-LIIF\cite{liif}} & $\times 2$$\times 3$$\times 4$ & 36.77 & 34.38 & 32.53 & 31.31 & 30.32 & 29.65  \\
    & All & \textbf{37.14} & \textbf{34.41} & \textbf{32.60} & \textbf{31.40} & \textbf{30.34} & \textbf{29.70}  \\
    \midrule
    \multirow{2}{*}{RDN-LTE\cite{lte}} & $\times 2$$\times 3$$\times 4$ & 36.84 & 34.44 & 32.63 & 31.43 & 30.44 & 29.76  \\
    & All & \textbf{37.24} & \textbf{34.52} & \textbf{32.76} & \textbf{31.53} & \textbf{30.54} & \textbf{29.84}  \\\midrule
    \multirow{2}{*}{RDN-CiaoSR\cite{ciaosr}} & $\times 2$$\times 3$$\times 4$ & 36.84 & 34.68 & 32.91 & 31.65 & 30.67 & 29.95  \\
    & All & \textbf{37.38} & \textbf{34.70} & \textbf{32.96} & \textbf{31.68} & \textbf{30.77} & \textbf{30.07}  \\\midrule
    \multirow{2}{*}{RDN-DDIR} & $\times 2$$\times 3$$\times 4$ & 37.22 & 34.81 & 32.99 & 31.76 & 30.77 & 30.05  \\
    & All & \textbf{37.63} & \textbf{35.02} & \textbf{33.20} & \textbf{31.91} & \textbf{30.94} & \textbf{30.21}  \\
    \bottomrule
  \end{tabular}
\end{table*}

\begin{table*}[h!]
  \caption{Quantitative Analysis of training scale factors in RealArbiSR dataset. The highest PSNR at each scale factor on each method is bolded. `$\times 2$$\times 3$$\times 4$' represents the models are trained at the scale factors of $\times 2.0$, $\times 3.0$, and $\times 4.0$. `All' represents the models are trained at the scale factors of  $\times 1.5$, $\times 2.0$, $\times 2.5$, $\times 3.0$, $\times 3.5$, and $\times 4.0$. The models are tested at the scale factors of $\times 1.7$, $\times 2.3$, $\times 2.7$, $\times 3.3$, and $\times 3.7$ in RealArbiSR dataset. }
  \label{tableS3}
  \centering
  \begin{tabular}{c | c|ccccc }
    \toprule
     Method & Training Scale &  $\times 1.7$ & $\times 2.3$ & $\times 2.7$ & $\times 3.3$ & $\times 3.7$  \\
    \midrule
     \multirow{2}{*}{RDN-LIIF\cite{liif}} & $\times 2$$\times 3$$\times 4$ & 34.63 & 32.24 & 31.31 & 30.18 & 29.63   \\
     & All & \textbf{34.66} & \textbf{32.40} & \textbf{31.45} & \textbf{30.28} & \textbf{29.71}   \\
    \midrule
    \multirow{2}{*}{RDN-LTE\cite{lte}} & $\times 2$$\times 3$$\times 4$ & 34.71 & 32.40 & 31.47 & 30.32 & 29.74 \\
     & All & \textbf{34.74} & \textbf{32.44} & \textbf{31.55} & \textbf{30.39} & \textbf{29.81}   \\\midrule
    \multirow{2}{*}{RDN-CiaoSR\cite{ciaosr}} & $\times 2$$\times 3$$\times 4$ & 34.45 & 32.46 & 31.66 & 30.55 & 29.94   \\
     & All & \textbf{34.54} & \textbf{32.50} & \textbf{31.67} & \textbf{30.56} & \textbf{29.96}  \\\midrule
    \multirow{2}{*}{RDN-DDIR} & $\times 2$$\times 3$$\times 4$ & 35.06 & 32.75 & 31.84 & 30.70 & 30.12   \\
     & All & \textbf{35.07} & \textbf{32.88} & \textbf{31.96} & \textbf{30.75} & \textbf{30.15}   \\
    \bottomrule
  \end{tabular}
\end{table*}

\section{Out-of-distribution Testing in RealArbiSR Dataset}
We conduct out-of-distribution testing in RealArbiSR dataset. To do this, we train one model at the scale factors of $\times 1.5$ $\times 2.0$ $\times 2.5$ $\times 3.0$, and $\times 3.5$, and test it at the scale factors of $\times 3.7$ and $\times 4.0$ in RealArbiSR dataset. As shown in Table \ref{tableS4}, our DDIR model achieves the best results in the out-of-distribution testing, compared to other baselines. 

\begin{table}
  \caption{Quantitative comparison of out-of-distribution testing on RealArbiSR dataset in PSNR(dB). The highest PSNR at each scale factor is bolded. One model is trained at the scale factors of $\times 1.5$, $\times 2.0$, $\times 2.5$, $\times 3.0$, and $\times 3.5$, and tested at the scale factors of $\times 3.7$ and $\times 4.0$ in RealArbiSR dataset.}
  \label{tableS4}
  \centering
  \begin{tabular}{ l |cc | cc}
    \toprule
    \multirow{2}{*}{Method} & \multicolumn{2}{c|}{EDSR Backbone} & \multicolumn{2}{c}{RDN Backbone} \\ 
    & $\times 3.7$ & $\times 4.0$ & $\times 3.7$ & $\times 4.0$ \\
    \midrule
    
     LIIF \cite{liif} & 29.59 & 29.49 & 29.73 & 29.72    \\

     LTE\cite{lte} & 29.55 & 29.64 & 29.83 & 29.92  \\

     CiaoSR\cite{ciaosr} & 29.84 & 29.86 & 30.03 & 29.99  \\
     DDIR & \textbf{29.88} & \textbf{29.99} & \textbf{30.11} & \textbf{30.11} \\

    \bottomrule
  \end{tabular}
\end{table}

\section{Analysis on Simulated and Real SR Experiments in RealArbiSR Dataset}
We compare the bicubic and real-world degradation in RealArbiSR dataset. It demonstrates the advantage of our RealArbiSR dataset compared to synthetic scale arbitrary methods with bicubic degradation. For our DDIR model with bicubic degradation, we remove the deformation field and deformation branch because they are specifically designed for real-world degradation and do not work for bicubic degradation. As shown in Table \ref{tableS5}, the performance of models with bicubic degradation all drops by a large margin. It proves bicubic degradation in synthetic scale arbitrary super-resolution fails to generalize in real-world degradation for all models at all scale factors. Further qualitative comparison is shown in Figure \ref{figS5} and \ref{figS6}. 

\begin{table*}
  \caption{Quantitative Analysis of bicubic and real-world degradations with the RDN \cite{rdn} backbone in RealArbiSR dataset. The highest PSNR at each scale factor on each method is bolded. The models are trained and tested at the scale factors from $\times 1.5$ to $\times 4.0$ with a step of $\times 0.5$ in RealArbiSR dataset. }
  \label{tableS5}
  \centering
  \begin{tabular}{c | c|cccccc }
    \toprule
     Method & Degradation &  $\times 1.5$ & $\times 2.0$ & $\times 2.5$ & $\times 3.0$ & $\times 3.5$ & $\times 4.0$ \\
    \midrule
    \multirow{2}{*}{RDN-LIIF\cite{liif}} & Bicubic & 35.70 & 32.69 & 30.91 & 29.63 & 28.71 & 28.03  \\
     & Real &  \textbf{37.14} & \textbf{34.41} & \textbf{32.60} & \textbf{31.40} & \textbf{30.34} & \textbf{29.70} \\
    \midrule
    \multirow{2}{*}{RDN-LTE\cite{lte}} & Bicubic & 35.68 & 32.69 & 30.89 & 29.61 & 28.69 & 28.01 \\
     & Real & \textbf{37.24} & \textbf{34.52} & \textbf{32.76} & \textbf{31.53} & \textbf{30.54} & \textbf{29.84}  \\\midrule
    \multirow{2}{*}{RDN-CiaoSR\cite{ciaosr}} & Bicubic & 35.67 & 32.69 & 30.89 & 29.62 & 28.71 & 28.03  \\
     & Real & \textbf{37.38} & \textbf{34.70} & \textbf{32.96} & \textbf{31.68} & \textbf{30.77} & \textbf{30.07}  \\\midrule
    \multirow{2}{*}{RDN-DDIR} & Bicubic & 35.70 & 32.70 & 30.91 & 29.64 & 28.72 & 28.04  \\
    & Real & \textbf{37.63} & \textbf{35.02} & \textbf{33.20} & \textbf{31.91} & \textbf{30.94} & \textbf{30.21} \\
    \bottomrule
  \end{tabular}
\end{table*}
\section{Quantitative Comparison between DDIR and Real-World SR Methods}
We compare the quantitative results between our DDIR model and other real-world SR methods \cite{realsr, CDC, d2c} on RealSR dataset in PSNR(dB). For our DDIR method, one model is trained and tested at the scale factors of $\times 2.0$, $\times 3.0$, and $\times 4.0$. For other real-world SR methods, different models are trained and tested for different scales. As shown in Table \ref{tableS6}, even with only one model, our DDIR model still outperforms existing real-world SR methods, at which one model is trained at each scale factor.  
\begin{table}[t]
  \caption{Quantitative comparison between DDIR and real-world SR methods on RealSR dataset in PSNR(dB). The best metric result at each scale factor is bolded. For DDIR, one model is trained and tested at the scale factors of $\times 2.0$, $\times 3.0$, and $\times 4.0$. LP-KPN\cite{realsr}, CDC\cite{CDC}, and D2C-SR\cite{d2c} train and test different models at different scales. }
  \label{tableS6}
  \centering
  \begin{tabular}{l |ccc }
    \toprule
    \multirow{2}{*}{Method} & \multicolumn{3}{c}{RealSR} \\ 
    & $\times2.0$ & $\times3.0$ & $\times4.0$ \\ 
    \midrule
    LP-KPN\cite{realsr} & - & 30.60 & 28.65   \\ 
    CDC\cite{CDC} & 33.96 & 30.99 & 29.24 \\ 
    RDN-D2C\cite{d2c} & 34.03 & 30.93 & 29.32 \\ 
    RDN-DDIR & \textbf{34.35} & \textbf{31.15} & \textbf{29.48} \\
    \bottomrule
  \end{tabular}
\end{table}

\section{More Visual Results}
\begin{figure*}[t!]
  \centering
  \includegraphics[width=1\linewidth]{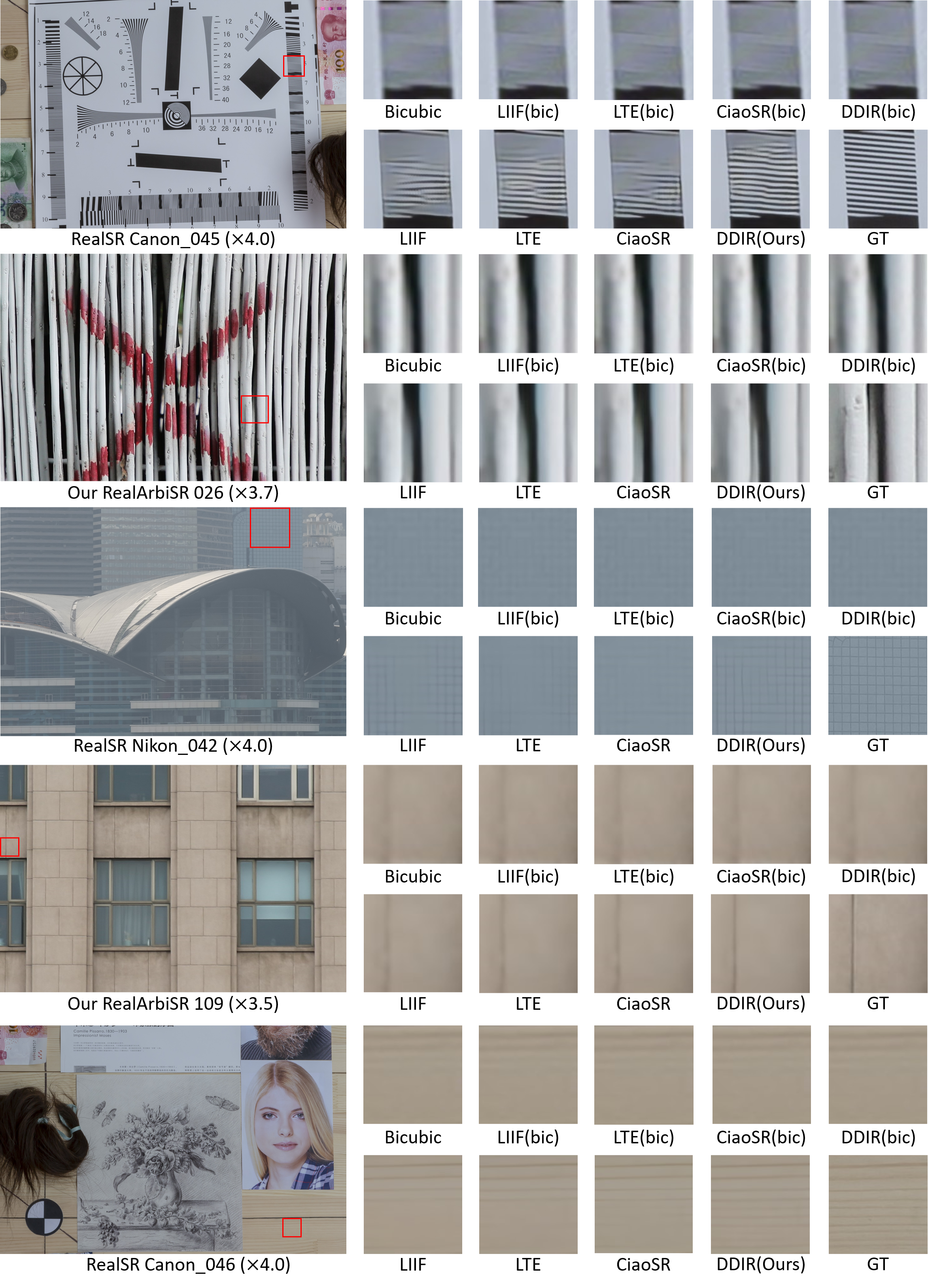}
  \caption{Qualitative comparisons between different methods on benchmarks. Zoom in to have better views.}
  \label{figS5}
\end{figure*}

\begin{figure*}[t!]
  \centering
  \includegraphics[width=1\linewidth]{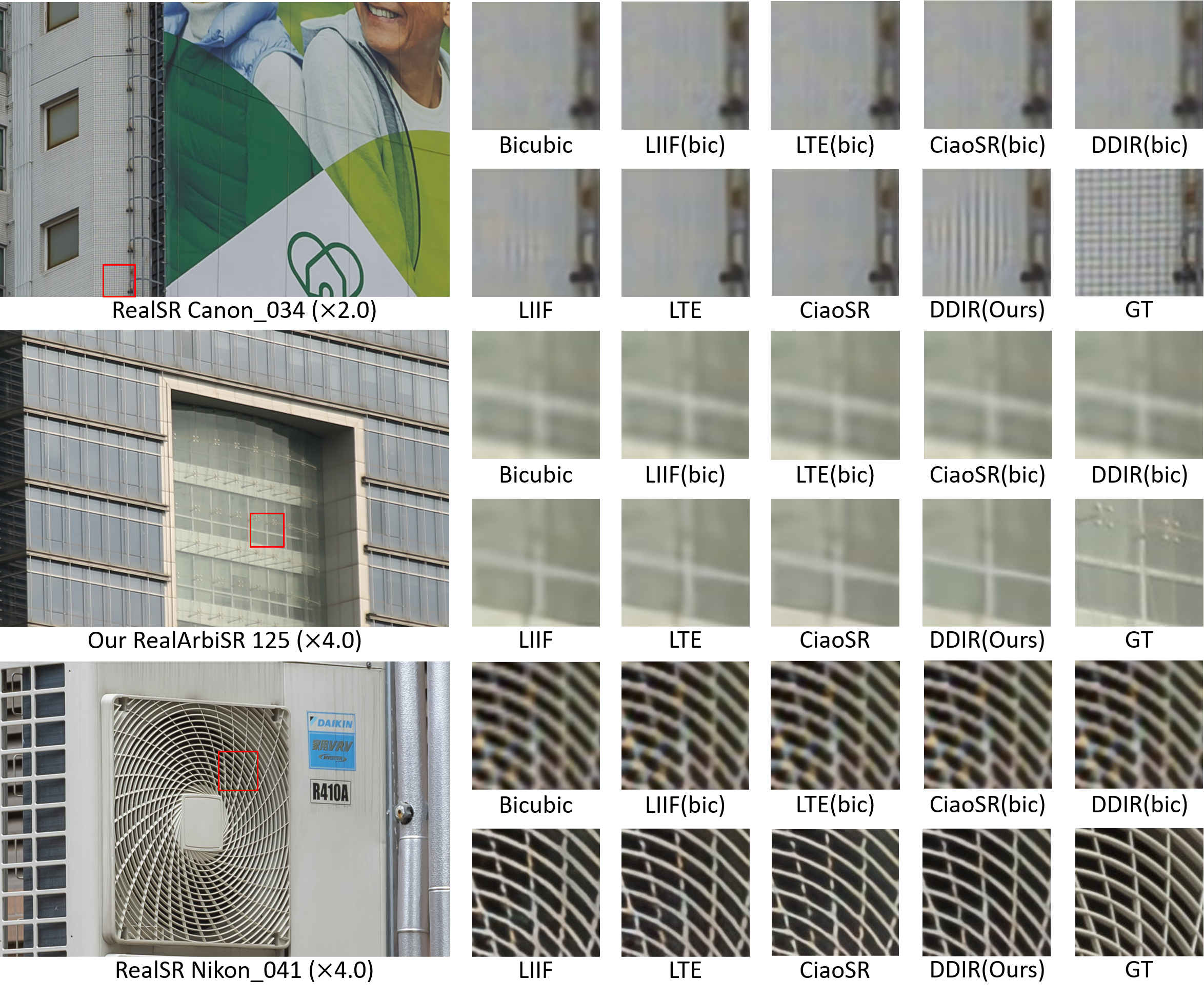}
  \caption{Qualitative comparisons on benchmarks. Zoom in to have better views.}
  \label{figS6}
\end{figure*}

We show more visual results on the RealArbiSR dataset and the RealSR dataset with real-world and bicubic degradations in Figure \ref{figS5} and \ref{figS6}. For DDIR model with bicubic degradation, we also remove the deformation branch and keep the appearance embedding. As shown in Figure \ref{figS5} and \ref{figS6}, our DDIR model reconstructs better image details and sharper edges compared to other methods. By comparing models with bicubic and real-world degradation, we can see synthetic scale arbitrary super-resolution methods with bicubic degradation fail to generalize in the real-world case.  

\end{document}